\documentclass{article}

\usepackage[preprint]{neurips_2025}

\usepackage[utf8]{inputenc} 
\usepackage[T1]{fontenc}    
\usepackage{hyperref}       
\usepackage{url}            
\usepackage{booktabs}       
\usepackage{amsfonts}       
\usepackage{nicefrac}       
\usepackage{microtype}      
\usepackage{xcolor}         
\usepackage{amsmath}
\usepackage{graphicx} 
\usepackage{multirow}
\usepackage{makecell}
\usepackage{siunitx}
\usepackage{subfigure}
\usepackage{graphicx}
\usepackage{amsmath}
\usepackage{gensymb}
\usepackage{mathrsfs}
\usepackage{multirow}
\usepackage{bbding}
\usepackage{wrapfig}
\usepackage{tabularx}
\usepackage{array}
\usepackage{subcaption}
\usepackage{booktabs}
\usepackage{amsmath}  
\usepackage{amsmath}

\title{Dendritic Resonate-and-Fire Neuron for Effective and Efficient Long Sequence Modeling}

\author{\textbf{Dehao Zhang}$^{1}$, \textbf{Malu Zhang}$^{1,}$\thanks{Corresponding author}, \textbf{ Shuai Wang}$^{1}$, \textbf{Jingya Wang}$^{1}$, \textbf{Wenjie Wei}$^{1}$, \textbf{Zeyu Ma}$^{1}$, \\ 
\textbf{Guoqing Wang}$^{1}$, \textbf{Yang Yang}$^{1}$, \textbf{Haizhou Li}$^{2}$ \\
  $^{1}$ University of Electronic Science and Technology of China\\
  $^{2}$ The Chinese University of Hong Kong (Shenzhen) \\
  \texttt{zhangdh@std.uestc.edu.cn, maluzhang@uestc.edu.cn} \\
}

\begin{document}

\maketitle
\begin{abstract}
The explosive growth in sequence length has intensified the demand for effective and efficient long sequence modeling. Benefiting from intrinsic oscillatory membrane dynamics, Resonate-and-Fire (RF) neurons can efficiently extract frequency components from input signals and encode them into spatiotemporal spike trains, making them well-suited for long sequence modeling. However, RF neurons exhibit limited effective memory capacity and a trade-off between energy efficiency and training speed on complex temporal tasks. Inspired by the dendritic structure of biological neurons, we propose a Dendritic Resonate-and-Fire (D-RF) model, which explicitly incorporates a multi-dendritic and soma architecture. Each dendritic branch encodes specific frequency bands by utilizing the intrinsic oscillatory dynamics of RF neurons, thereby collectively achieving comprehensive frequency representation. Furthermore, we introduce an adaptive threshold mechanism into the soma structure that adjusts the threshold based on historical spiking activity, reducing redundant spikes while maintaining training efficiency in long sequence tasks. Extensive experiments demonstrate that our method maintains competitive accuracy while substantially ensuring sparse spikes without compromising computational efficiency during training. These results underscore its potential as an effective and efficient solution for long sequence modeling on edge platforms.
\end{abstract}

\section{Introduction}
Long sequence modeling efficiently captures complex temporal patterns and dynamic characteristics, demonstrating exceptional application potential in edge computing scenarios such as speech recognition~\cite{luo2020dual, wang2025ternary, wang2024global} and electroencephalogram (EEG) monitoring~\cite{sayeed2019eseiz,wang2020accurate}. However, mainstream sequence modeling methods still primarily rely on Recurrent Neural Networks (RNNs)~\cite{schuster1997bidirectional}, Transformers~\cite{vaswani2017attention}, and state-space models (SSMs)~\cite{gu2021combining, gu2022parameterization}. Although these approaches effectively compress contextual information into finite states, they still involve extensive floating-point matrix multiplications, resulting in high computational complexity, inference latency, and energy consumption~\cite{shen2025spikingssms, bal2024p}. Therefore, designing long sequence models that simultaneously achieve high performance, energy efficiency, and fast inference remains an essential and ongoing challenge. 

Inspired by the structure and function of neural circuits in the brain, Spiking Neural Networks (SNNs) have emerged as a biologically plausible and computationally efficient model~\cite{maass1997networks,tavanaei2019deep}. Unlike Artificial Neural Networks (ANNs), SNNs possess event-driven computational capabilities~\cite{caviglia2014asynchronous, zhang2021rectified} and the potential to process dynamic temporal information~\cite{yin2021accurate, zheng2024temporal}. These properties allow information to be transmitted through binary spikes and enable the retention of historical context via membrane potentials. 
Several studies have applied SNNs for long sequence modeling based on Leaky Integrate-and-Fire (LIF) neurons. Zhang et al.~\cite{zhang2024tc} introduce a two-compartment LIF neuron to better capture long-term temporal dependencies, while Fang et al.~\cite{fang2024parallel} remove the reset mechanism to maximize the utilization of information across all timesteps. However, due to their serial charge-fire-reset dynamics, these methods fail to capture complex timescale variations and long-range dependencies in long sequences~\cite{yin2021accurate, bellec2018long, wang2025spiking}, limiting their performance in complex tasks.
As an efficient alternative to LIF neurons, Resonate-and-Fire (RF) neurons~\cite{izhikevich2001resonate} employ complex-valued state variables to more effectively retain historical information in long sequences modeling. Higuchi et al.~\cite{higuchibalanced} integrate a refractory mechanism into the reset dynamics of RF neurons, maintaining frequency selectivity while significantly improving long sequence processing capacity and promoting spike pattern sparsity. Huang et al.~\cite{huang2024prf} remove the reset mechanism and propose that a learnable time constant can capture the intrinsic reset behavior of RF neurons, thereby reducing the computational complexity from \(\mathcal{O}(\mathcal{L}^2)\) to \(\mathcal{O}(\mathcal{L} \log \mathcal{L})\), and significantly enhancing computational efficiency. Despite these advancements, current methods still face two major challenges. First, the limited bandwidth of RF neurons constrains their ability to extract diverse frequency-band combinations from complex temporal signals, 
making them behave more like simplified resonators. 
Second, RF neurons encounter a trade-off between energy efficiency and training speed: removing the reset mechanism enables efficient training but leads to excessive spike activity, whereas incorporating a reset suppresses spiking activity at the cost of increased training overhead.

Inspired by the dendritic structure of biological neurons~\cite{rall1994theoretical, traub1991model}, we propose a novel Dendritic Resonate-and-fire (D-RF) neuron for effective and efficient long sequence modeling. It consists of a multi-dendrite and soma structure. First, each dendritic branch captures specific frequency responses from input signals through the oscillation characteristics of RF neurons, thereby achieving comprehensive spectral decomposition across multiple timescales. Second, we incorporate an adaptive threshold mechanism into the soma structure that dynamically adjusts thresholds based on historical spiking patterns, achieving sparse spikes while maintaining training efficiency. Extensive experiments on long sequence tasks confirm the high-performance and energy-efficient of the proposed neuron model. The main contributions are summarized as follows:


\begin{itemize}
\item We conduct a detailed analysis of the limitations of existing RF neurons in long sequence modeling, highlighting their restricted memory capacity and the inherent trade-off between energy efficiency and training speed. First, the limited bandwidth response causes RF neurons to behave like simplified resonators. Second, the presence or absence of a reset mechanism leads to a conflict between sparse spiking and training efficiency.


\item 
We propose the D-RF neuron, which comprises two components. First, dendritic branches exploit RF dynamics to achieve specialized frequency selectivity, collectively enabling full spectral coverage across multiple timescales. Second, an adaptive threshold mechanism in the soma dynamically adjusts thresholds based on history spiking activity, balancing computational cost and energy efficiency while preserving training effectiveness.


\item Extensive experiments demonstrate that our method achieves competitive performance across various long sequence tasks. Moreover, the method produces sparser spiking activity while maintaining training efficiency. These results highlight our model's dual advantages in effectiveness and computational efficiency for long sequence modeling. 
\end{itemize}

\section{Related Work}
\label{sec:related}
\subsection{Advanced Spiking Neurons for Long Sequence Modeling}
Due to the dynamic characteristics of spiking neurons, it is believed that they have the ability to handle long sequence modeling. However, the LIF model~\cite{gerstner2002spiking, izhikevich2003simple} and its variants~\cite{bellec2018long, yin2021accurate, fang2021incorporating} exhibit limited memory capacity in temporal tasks, which is considered a critical factor for effective long sequence modeling. To overcome this limitation, several studies~\cite{chen2024pmsn, zhang2024tc} draw inspiration from more complex neural dynamics. RF neurons~\cite{izhikevich2001resonate} have attracted considerable attention due to their intrinsic frequency band preference. Orchard et al.~\cite{orchard2021efficient} leverage RF neurons to encode raw signals into sparse spike trains, thereby significantly reducing output bandwidth. Furthermore, Higuchi et al.~\cite{higuchibalanced} introduce adaptive decay factor mechanisms~\cite{shaban2021adaptive, ganguly2024spike} and refractory mechanisms~\cite{simoes2024thermodynamic}, improving the balance between energy efficiency and performance of RF neurons in long-range sequence modeling. In addition, RF-based models demonstrate competitive performance in sequence modeling tasks, including image classification~\cite{hille2022resonate}, optical flow tracking~\cite{frady2022efficient}, and audio processing~\cite{zhang2024spike, shrestha2024efficient}. Moreover, RF neurons can be efficiently implemented on neuromorphic hardware like the Loihi~\cite{davies2018loihi, orchard2021efficient}.


\subsection{Training Strategies for Spiking Neural Networks}
The mainstream training methods for deep SNNs can be categorized into ANN-to-SNN conversion~\cite{diehl2015fast, rueckauer2017conversion, wang2025training} and direct training~\cite{wu2018spatio, fang2021deep}. ANN-to-SNN conversion methods use the similarity between spike firing rates and ANN activation functions, but need many timesteps to reach high accuracy. In contrast, direct training enables SNNs to achieve performance comparable to ANNs with the same architecture within a limited number of timesteps. Specifically, direct training introduces surrogate gradient functions~\cite{deng2022temporal, neftci2019surrogate} to enable backpropagation, thereby addressing the non-differentiability of spike firing functions. However, applying direct training to long sequence tasks presents greater challenges~\cite{shen2025spikingssms}, as such tasks often require thousands of timesteps. Consequently, some research~\cite{meng2023towards, hu2024high, su2024snn, zhang2025memory} tend to explore more efficient training strategies for SNNs. 
Yin et al.\cite{yin2024understanding} further explore intrinsic challenges in SNNs training, especially the reset behavior of neurons. Therefore, Fang et al.~\cite{fang2024parallel} change the dynamic process of spiking neurons into a learnable matrix, avoiding the reset mechanism. Similarly, Shen et al.~\cite{shen2025spikingssms} introduce the SDN block to simulate the reset process. These methods greatly reduce the training cost of SNNs while keeping their asynchronous inference ability.

\section{Preliminary}
\label{sec:prelim}
\subsection{Resonate-and-Fire Neuron}
Inspired by the damped and sustained subthreshold oscillations observed in the membrane potentials of mammalian nervous systems~\cite{alonso1989subthreshold, llinas1991vitro, pedroarena1997dendritic}, RF neurons are proposed~\cite{izhikevich2001resonate}. Given input signal \(\mathcal{I}(t)\), the dynamics of the RF neuron at timestep \(t\) can be described as follows:
\begin{equation}
\frac{d}{dt} z(t) = (b + \text{i} \omega) z(t) + \mathcal{I}(t), \label{1}
\end{equation}
\( z = u + \text{i}v \in \mathbb{C} \) represents the complex state of RF neurons, where \(u\) represents a current‐like variable capturing voltage‐gated and synaptic current dynamics, and the imaginary component \(v\) corresponds to a voltage‐like variable. \( \omega > 0 \) denotes the angular frequency of the neuron, indicating the number of radians it oscillates per second, while the damping factor \( b < 0 \) regulates the exponential decay of the oscillation. It can be discretized using the Euler method~\cite{atkinson2008introduction}:
\begin{equation}
z[t] = \exp \left\{ \delta (b + \mathrm{i} \omega) \right\} \cdot z[t-1] + \delta \mathcal{I}[t],\label{2}
\end{equation}
\( \delta \) is the discrete timestep. When the real part of \(z[t]\) exceeds the threshold, the neuron fires a spike; otherwise, it remains silent. 
Additionally, RF neuron exhibits a preference for specific frequency bands. As shown in Fig.~\ref{problem}(a), we present the oscillatory behavior under spike inputs with different frequencies. It is observed that both the membrane potential and the phase state accumulate rapidly.

\subsection{Direct Training in Spiking Neural Networks}
Due to the BPTT~\cite{wu2018spatio} and Surrogate Gradient methods~\cite{neftci2019surrogate}, training large-scale SNNs becomes feasible. Specifically, the gradient of the weight \( w^l \) at timestep \( T \) can be represented as follows:
\begin{equation}
    \nabla_{w^l} \mathcal{L} = \sum_{l=1}^{T} \left( \frac{\partial \mathcal{L}}{\partial u^l(t)} \right) S^{l-1}(t), \label{3}
\end{equation}
where $u^l(t)$ and $S^l(t)$ denote the membrane potential and spike emission of the $l$-th layer at time $t$, respectively. \( \mathcal{L} \) is the loss function. It can be calculated as follows:
\begin{equation}
\frac{\partial \mathcal{L}}{\partial u^l(t)} = \frac{\partial \mathcal{L}}{\partial S^l(t)} \frac{\partial S^l(t)}{\partial u^l(t)} + \frac{\partial \mathcal{L}}{\partial u^l(t+1)} \frac{\partial u^l (t+1)}{\partial u^l(t)}, \label{4}
\end{equation}
\begin{equation}
\frac{\partial \mathcal{L}}{\partial S^l(t)} = \frac{\partial \mathcal{L}}{\partial u^{l+1}(t)} \frac{\partial u^{l+1}(t)}{\partial S^l(t)} + \frac{\partial \mathcal{L}}{\partial u^l(t+1)} \frac{\partial u^l (t+1)}{\partial S^l(t)}. \label{5}
\end{equation}
The non-differentiable term \(\frac{\partial S^l (t)}{\partial u^l(t)}\) can be substituted with a surrogate function. \( \frac{\partial u^l (t+1)}{\partial S^l(t)}\) and \(\frac{\partial u^l (t+1)}{\partial u^l(t)}\) are the temporal gradients that need to be calculated. As shown in Eq.~\ref{4} and Eq.~\ref{5}, the gradient at each timestep depends not only on the current states but also recursively on future states. This recursive dependency across layers and timesteps results in a computational complexity of \(\mathcal{O} (\mathcal{L}^2)\). 

\section{Method}
\subsection{Problem Analysis}
\begin{figure}[htpb] 
\centering
\includegraphics[width=1.0\linewidth]{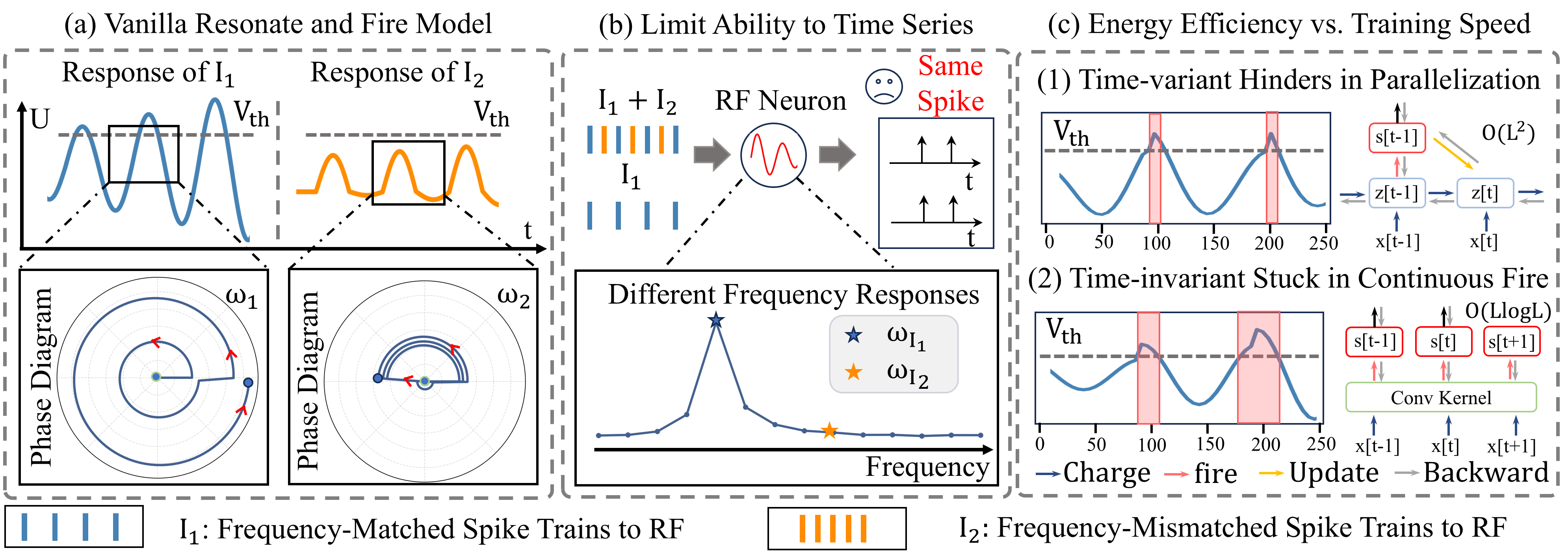}
\caption{Problem Analysis: (a) Response of Different Spike Trains: Frequency-matched inputs lead to rapid membrane potential accumulation, whereas mismatched inputs yield weaker responses. 
(b) Limited Ability to Time Series: 
A single RF neuron struggles to respond to different frequency-varying inputs due to its narrow band selectivity. (c) Energy Efficiency vs. Training Speed: The time-variant method enables sparse spiking but has a high training cost of $\mathcal{O}(\mathcal{L}^2)$. The time-invariant method allows faster training with complexity of $\mathcal{O}(\mathcal{L} \log \mathcal{L})$, but often leads to continuous fire.
}
\label{problem}
\end{figure}

\textbf{Limited Performance in Complex Tasks}: Benefiting from the design of the decay kernel and complex-valued states, RF neurons exhibit pronounced frequency selectivity. However, this property also limits the model’s ability to discriminate diverse input patterns. As shown in Fig.~\ref{problem}(b), input signals with distinct frequency components may elicit similar spiking responses, as components misaligned with the neuron’s intrinsic frequency tend to be suppressed, thereby impairing the network’s capacity to capture and distinguish diverse temporal features. To further verify this limitation, we visualize the frequency response of a single RF neuron. The results show that the neuron responds primarily within a narrow bandwidth and reaches its peak at the intrinsic frequency, making it difficult to capture complex frequency compositions. This observation highlights an inherent representational limitation of RF neurons in modeling complex temporal features.

\textbf{Challenges in Balancing Energy Efficiency and Training Speed}: In long sequence tasks, RF neurons incur significant training costs and potentially excessive energy consumption. Existing research focuses on two main strategies. First, as shown in Fig.~\ref{problem}(c).1, Higuchi et al.~\cite{higuchibalanced} propose an adaptive decay factor that instantly raises the decay coefficient after each spike, suppressing membrane potential oscillations and promoting sparse spikes. However, this approach requires temporal unfolding during backpropagation, resulting in $\mathcal{O}(\mathcal{L}^2)$ complexity. Second, Huang et al.~\cite{huang2024prf} employ a learnable decay factor to the reset mechanism and use a Fourier transform reformulation to reformulate the temporal dynamics into a parallel convolutional process between an oscillatory kernel and the input signal, reducing computational complexity to $\mathcal{O}(\mathcal{L}\log\mathcal{L})$. Nevertheless, due to the decay factor remains time‐invariant, the model preserves its pre‐spike amplitude. As shown in Fig.~\ref{problem}(c).2, it results in sustained burst firing that undermines the low‐power advantages of SNNs.

\subsection{Dendritic Resonate-and-Fire Neuron}
\label{Dendritic}
To better capture features across different frequency bands, we propose the D-RF neuron model. Unlike the vanilla RF model~\cite{izhikevich2001resonate}, D-RF model comprises a soma and multiple dendritic branches. Each dendritic branch extracts state responses corresponding to specific frequency preferences in the input signal $\mathcal{I}[t]$. When $\mathcal{I}[t]$ contains frequency components aligned with a branch’s preference, the membrane potential of that branch accumulates rapidly. The soma integrates input currents from all dendritic branches and generates a spike once its membrane potential exceeds a predefined threshold. Specifically, the membrane potential dynamics of the $i$-th dendritic branch is defined as follows:
\begin{equation}
    \frac{dz_i(t)}{dt} = \{-1 / \tau_i + \mathrm{i} \omega_i\} \cdot z_i(t) + \gamma_i \mathcal{I}(t), \label{6}
\end{equation}
where \( \tau_i \) and \( \omega_i \) represent the decay factor and membrane potential oscillation coefficient associated with the \( i \)-th dendritic branch, respectively. \( \mathcal{I}(t) \) denotes the presynaptic input at time \( t \), and \( \gamma_i \) represents the membrane capacitance of the \( i \)-th dendritic branch. This modeling framework allows different dendritic branches to selectively respond to specific frequency components. To enable efficient inference, the Zero-Order Hold (ZOH) method~\cite{decarlo1989linear} is employed for discretization. The membrane potential dynamics of all dendrites are given by:
\begin{equation}
\mathcal{Z}[t] = \exp\left\{
\begin{bmatrix}
-\frac{1}{\tau_1} + \text{i}\omega_1 & 0 & \cdots & 0 \\
0 & -\frac{1}{\tau_2} + \text{i}\omega_2 & \cdots & 0 \\
\vdots & \vdots & \ddots & \vdots \\
0 & 0 & \cdots & -\frac{1}{\tau_n} + \text{i}\omega_n
\end{bmatrix} \cdot \delta
\right\} \mathcal{Z}[t-1] + \Gamma^l I[t]. \label{7}
\end{equation}
Here, $\delta$ denotes the discrete timestep, and $\mathcal{Z} = [z_1, z_2, \cdots, z_n]^T$ represents the states of individual dendritic branches, with $\Gamma = [\gamma_1, \gamma_2, \cdots, \gamma_n]^T$ denoting their respective time constants. To further enhance the frequency characteristics across dendritic branches, each branch is assigned an individual importance weight. The dynamics of soma are defined as follows:
\begin{equation}
H[t] = \mathcal{C} \Re\{Z(t)\}, \quad \quad S[t] = \Theta\left(H[t] - V_{th}[t]\right), \label{8}
\end{equation}
$\mathcal{C} \in \mathbb{R}^{n \times 1}$ denotes the importance weights assigned to each dendritic branch. $\Theta(\cdot)$ represents the Heaviside function. When the presynaptic membrane potential of the soma \(H[t]\) exceeds the threshold \(V_{\text{th}}\), a spike \(S[t]\) is generated.

We further analyze the effectiveness of the dendrite design by examining its frequency band responses. For time-invariant RF neuron with decay kernel $b + \text{i}\omega$, it can be modeled as a time-invariant convolutional process with kernel $h(n)$, defined as $h(n) =\exp\{\delta (b + \text{i}\omega)\}^n$. 
Its frequency response is described as follows:
\begin{equation}
        \left|\left|H(\exp\{\text{i}\Omega\})\right|\right| = \left|\left|\sum_{n=0}^{\infty}h(n) \exp\{-\text{i}\Omega n\}\right|\right|=\left|\left|\frac{\delta}{1-\exp\{\delta b + \text{i} (\delta \omega - \Omega)\}}\right|\right| \label{9}.
\end{equation}
Therefore, a single RF neuron can be regarded as a first-order band-pass filter with a resonance peak at $\Omega \approx \omega$, and a narrow frequency band determined by the damping factor $\delta b$ (as shown in Fig.~\ref{problem}(b)). In contrast, the frequency response of our D-RF model is defined as follows:
\begin{equation}
        \left|\left|H_{\text{D-RF}}(\exp\{\text{i}\Omega\})\right|\right| =\sum_{i=1}^{n} \mathcal{C}_i \cdot \left|\left|H(\exp\{\text{i}\Omega\})\right|\right|,\quad B_{\text{eff}} \approx \sum_{i=1}^{n} \beta_i \left|\left|\frac{\tau_i}{\delta }\right|\right|, \label{10}
\end{equation}
$B_{\text{eff}}$ denotes the total frequency response of the D-RF neuron, while $\beta_i \in [0, 1)$ quantifies the independent contribution of the $i$-th branch to the overall frequency coverage. Consequently, compared to its single-branch counterpart, the D-RF neuron provides a significantly broader frequency sensitivity. 

\begin{figure}[htpb] 
\centering
\includegraphics[width=1.0\linewidth]{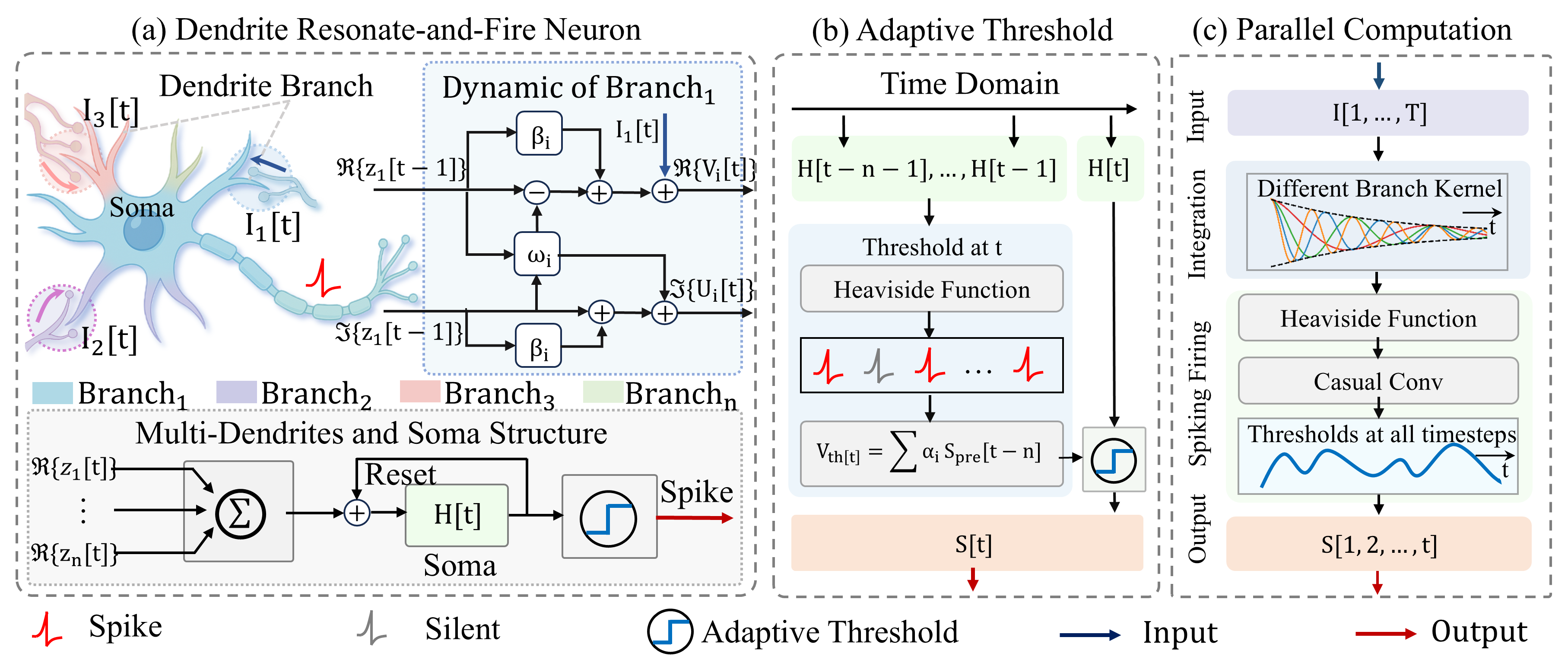}
\caption{The Structure of D-RF model: (a) Dendrite Resonate-and-Fire Neuron: D-RF model consists of multiple dendritic branches, each encoding frequency-specific dynamics. The soma integrates membrane potentials from all branches through an adaptive threshold mechanism to enable sparse spikes. (b) Adaptive Threshold: The threshold at time t is dynamically determined based on the historical spiking activity. (c) Parallel Computation: Different convolution kernels on each branch and a causal convolution for the adaptive threshold allow parallel processing of the input over time.
}
\label{method}
\end{figure}

\subsection{Adaptive Threshold for Accelerated and Efficient Learning}
\label{Adaptive}
To balance training speed and energy efficiency, we propose an adaptive thresholding strategy that dynamically adjusts the threshold based on the spiking activity from previous timesteps. Specifically, the threshold at timestep \(t\) is defined as:
\begin{equation}
V_{\text{th}}[t] =\sum_{k=1}^{n} \alpha_k \Theta(\Re\{ \mathcal{Z}[t-k-1, ..., t-1]\} - V_{\text{pre}}) + V_{\text{pre}},\label{11}
\end{equation}
where \( V_{\text{pre}} \) denotes the origin threshold (set to 1), and \(\alpha_k \in (0, 1)\) represents the importance of preceding spikes. As shown in Fig.~\ref{method}(b), $\alpha_k$ is the shared parameter during the adaptive threshold computation. This process can be interpreted as a one-dimensional causal convolution with kernel size $n$, where the kernel is defined as $\mathcal{A} = [\alpha_1, \dots, \alpha_n]$. The spiking process can be reformulated as: 
\begin{equation}
    \mathcal{S}[t] = \Theta\{\underbrace{C^l \Re\{\mathcal{Z}\}}_{\text{Dendritic Input}} - \underbrace{(\text{Conv1d}(\Theta (C^l \Re\{\mathcal{Z}\} - V_{\text{pre}}) + V_{\text{pre}})}_{\text{Adaptive Threshold}}\}[t]. \label{12}
\end{equation}
\(\text{Conv1d}(\cdot)\) present causal convolution along the temporal domain, enabling sparser spike activity while preserving the parallelizable nature of training. We demonstrate the effectiveness of this strategy by analyzing both the forward and backward propagation processes.

In the forward propagation, the computational complexity of the D-RF neuron is determined by multi-dendritic input and an adaptive threshold mechanism. Without the need for temporal unfolding, the D-RF neuron can be formulated in a parallelized manner. For the component of dendritic input, each dendritic branch functions independently and is decoupled in time. Therefore, the membrane potentials of all branches at time t are defined as follows:
\begin{equation}
\mathcal{Z}^l[t] = \textstyle \sum_{k=0}^{t} \Gamma^l \exp\left\{ k \cdot \delta \mathcal{D} \right\} \cdot \mathcal{I}^l[t - k], \label{13}
\end{equation}
$\mathcal{D}$ characterizes the oscillatory resonators of individual dendritic branches and is defined as a diagonal matrix: \( \mathcal{D} = \text{Diag}\left\{ -1/\tau_1  + \omega_1,\ -1/\tau_2 + \omega_2,\ \cdots,\ -1/\tau_n + \omega_n \right\} \). Since $\mathcal{D}$ is time-invariant, Eq.~\ref{13} can be reformulated as a convolution between the input signal $\mathcal{I}$ and the kernel $\mathcal{K}$:
\begin{equation}
\mathcal{Z}[t] = (\mathcal{K} * \mathcal{I})[t] = 
\mathcal{F}^{-1} \left\{ \mathcal{F}\{\mathcal{K}\} \cdot \mathcal{F}\{\mathcal{I}\} \right\}[t], \quad \mathcal{K} = \big[\, \delta \mathcal{D}^1,\ \delta \mathcal{D}^2,\ \dots,\ \delta \mathcal{D}^n \,\big], \label{14}
\end{equation}
$\mathcal{F}(\cdot)$ and $\mathcal{F}^{-1}(\cdot)$ denote the forward and inverse Fourier transform operations. Consequently, the charging process ensures a computational complexity of $\mathcal{O}(\mathcal{L} \log \mathcal{L})$ without introducing additional training overhead. For the adaptive threshold mechanism, its implementation via convolution operations is well-suited for GPU acceleration. Additionally, since $\alpha_k > 0$, a greater number of spikes in the previous timestep increases the threshold at time t, thereby promoting sparser spikes.

In the backward propagation stage, the adaptive threshold removes temporal dependencies between the gradient, enabling highly parallelizable training. Given $\mathcal{I}[t] = w^l S^{l-1}[t]$, the gradient of the loss with respect to the weight $w$ can be expressed as:
\begin{equation}
\nabla_{w^l}\mathcal{L} = \sum_{t=0}^{T} \underbrace{\frac{\partial \mathcal{L}}{\partial \mathcal{S}^{l}[t]} \frac{\partial \mathcal{S}^{l}[t]}{\partial \mathcal{Z}^{l}[t]} \frac{\partial \mathcal{Z}^{l}[t]}{\partial w^{l}} \vphantom{\frac{\partial^2}{\partial^2}}}_{\text{Sequential Training}} = \left\langle \underbrace{\frac{\partial \mathcal{L}}{\partial \mathcal{S}^{l}[t]} \frac{\partial \mathcal{S}^{l}[t]}{\partial \mathcal{Z}^{l}[t]} \vphantom{\frac{\partial^2}{\partial^2}}, (\mathcal{K} * \mathcal{S}^{l-1})[t] \vphantom{\frac{\partial^2}{\partial^2}}}_{\text{Parallel Training}} \right\rangle,\label{15}
\end{equation}
\(\langle \cdot, \cdot\rangle \) is the inner product.  
The derivative \(\frac{\partial \mathcal{S}^{l}[t]}{\partial \mathcal{Z}^{l}[t]}\) can be further defined as follows:
\begin{equation}
\frac{\partial \mathcal{S}[t]}{\partial \mathcal{Z}[t]} = \mathcal{C}^l \mathcal{G}\left(C^l\Re\{Z[t]\} - V_{\text{th}}[t]\right) \frac{\partial \Re\{\mathcal{Z}[t]\}}{\partial \mathcal{Z}[t]},\label{16}
\end{equation}
$\mathcal{G}(\cdot)$ denotes the surrogate gradient function. In this work, it is implemented as a double Gaussian function~\cite{yin2021accurate}. As shown in Eq.~\ref{15} and Eq.~\ref{16}, the gradient during backpropagation depends only on the current timestep. Consequently, the incorporation of the adaptive threshold introduces no additional training complexity, while encouraging sparse spike activity and preserving low computational cost.

\section{Experiment}
\label{sec:exper}
\subsection{Compare with the SOTA methods}
To validate the effectiveness of our proposed method, we conduct experiments on multiple time-series datasets. All experiments are conducted at least five times. First, we compare the performance of D-RF with other SOTA models on commonly datasets, including Spiking Heidelberg Digits (SHD)~\cite{cramer2020heidelberg} with 250 timesteps, Sequential MNIST, Permuted Sequential MNIST (S/PS-MNIST)~\cite{le2015simple} with 784 timesteps, and the more challenging Sequential CIFAR10 (S-CIFAR10)~\cite{chen2024pmsn} with 1024 timesteps. As presented in Table~\ref{table:performance_comparison_l1}, D-RF achieves SOTA performance while using fewer or comparable parameters. 

\begin{table}[!htpb]
\centering
\caption{Performance Comparison of Various Models.}
\resizebox{\textwidth}{!}{
\fontsize{9}{11}\selectfont  
\begin{tabular}{ccccccc}

\toprule
\textbf{Dataset}   & \textbf{Method} & \textbf{Model Size} & \textbf{Type} & \textbf{Parallel} & \textbf{Dendritic} & \textbf{Acc.} \\ \midrule
\multirow{10}{*}{ \begin{tabular}{@{}c@{}} S/PS-MNIST \\  (784 Timesteps) \end{tabular}} 
& LIF~\cite{zhang2024tc}    &    85.1K & FF       & \XSolidBrush       &    \XSolidBrush       &  72.06 / 10.00    \\
                            & ALIF~\cite{yin2021accurate}   &   156.3K   &  Rec   &   \XSolidBrush     &    \XSolidBrush       &  98.70 / 94.30    \\
                            & BRF~\cite{higuchibalanced}    &  68.9K & Rec       &    \XSolidBrush    &     \XSolidBrush      &   99.10 / 95.20   \\ 
                            & PSN~\cite{fang2024parallel}    &  2.5M & FF       &    \XSolidBrush    &     \Checkmark      &   97.90 / 97.80   \\ \cmidrule{2-7}                             
                            & TC-LIF~\cite{zhang2024tc} &   155.1K         &     Rec   &    \XSolidBrush       &  \Checkmark &  99.20 / 95.36   \\
                            & DH-LIF~\cite{zheng2024temporal} & 0.8M          &    Rec    &   \XSolidBrush  & \Checkmark      &   98.9 / 94.52   \\
                            & PMSN~\cite{chen2024pmsn}   &     156.4K       &   FF     &     \Checkmark      & \Checkmark &   99.50 / 97.80   \\
                            & \textbf{Ours}   &   155.1K    & FF     &  \Checkmark      &  \Checkmark         & 
                            \textbf{99.50} / \textbf{98.20} \\ \midrule
\multirow{10}{*}{ \begin{tabular}{@{}c@{}} SHD \\  (250 Timesteps) \end{tabular}} & LIF~\cite{zenke2021remarkable}    &    249.0K & Rec       & \XSolidBrush       &    \XSolidBrush       &  84.00   \\
                            & ALIF~\cite{yin2021accurate}   &    141.3K   &  Rec   &   \XSolidBrush     &    \XSolidBrush       &  84.40    \\
                            & BRF~\cite{higuchibalanced}    &  108.8K & Rec       &    \XSolidBrush    &     \XSolidBrush      &   92.50   \\ 
                            & PSN~\cite{fang2024parallel}    &  232.5K & FF       &    \Checkmark    &     \XSolidBrush      &    89.75   \\ \cmidrule{2-7}                             
                            & TC-LIF~\cite{zhang2024tc} &   141.8K         &     Rec   &    \XSolidBrush       &  \Checkmark &  88.91   \\
                            & DH-LIF~\cite{zheng2024temporal} & 0.5M          &    Rec    &   \XSolidBrush  & \Checkmark      &   91.34  \\
                            & PMSN~\cite{chen2024pmsn}   &      199.3K       &   FF     &     \Checkmark      & \Checkmark &    95.10   \\
                            & \textbf{Ours}   &   155.1K    & FF     &  \Checkmark      &  \Checkmark         &   \textbf{96.20} \\ \midrule
\multirow{7}{*}{ \begin{tabular}{@{}c@{}} S-CIFAR10 \\  (1024 Timesteps) \end{tabular}} & LIF~\cite{zenke2021remarkable}    &    0.18M  &  FF  & \XSolidBrush       &    \XSolidBrush       &  45.07  \\
                            & PSN~\cite{fang2024parallel}   &  6.47M   &  FF   &   \Checkmark     &    \Checkmark       &   55.24    \\
                            & SPSN~\cite{fang2024parallel}    &   0.18M & FF       &    \Checkmark    &     \Checkmark      &   70.23   \\ \cmidrule{2-7}                             
                            & PMSN~\cite{chen2024pmsn}   &      0.21M       &   FF     &     \Checkmark      & \Checkmark &    82.14   \\
                            & \textbf{Ours}   &   0.21M   & FF     &  \Checkmark      &  \Checkmark  &  \textbf{84.30} \\ \bottomrule
\end{tabular}
}
\label{table:performance_comparison_l1}
\end{table}

Additionally, experimental results demonstrate that the RF-based model consistently outperforms the similarly sized LIF model across all datasets. It further confirms the temporal modeling capabilities of RF neurons in long sequence tasks. Moreover, our model also demonstrates superior recognition performance compared to other dendritic models. Compared to the DH-LIF model~\cite{zheng2024temporal}, our approach enables parallel computation, effectively lowering the training speed associated with dendritic neurons. Compared to the PMSN model~\cite{chen2024pmsn}, our threshold resetting strategy prevents frequent spike generation and more effectively captures temporal dependencies. On the more challenging Sequential CIFAR10 dataset, our method achieves a recognition accuracy of 84.20\%, representing a 2.16\% improvement.

We also evaluate our approach on the more challenging LRA benchmark~\cite{tay2020long}. As shown in Table~\ref{table:performance_comparison_lra}, our model achieves significantly higher recognition accuracy than other neural models. Notably, it achieves 60.02\% accuracy on the ListOps task. It also demonstrates strong performance on tasks with longer timesteps, reaching 86.52\% precision on the Text task (4096 timesteps) and 90.02\% precision on the Retrieval task (4000 timesteps). 
For the Image Task, D-RF attains 85.32\% accuracy which underperforming SpikingSSM~\cite{shen2025spikingssms}. This gap stems from the use of LayerNorm in SpikingSSMs, which reduces temporal variance. Furthermore, the performance gap between our model and ANN-based approaches like S4~\cite{gu2021efficiently} is no greater than 3\%. Specifically, our model’s accuracy is 0.42\% higher for ListOps tasks. These findings highlight the strong temporal modeling ability of our method. 
\begin{table}[ht]
\centering
\caption{Comparison of Model Accuracy on LRA Benchmark.}
\resizebox{\textwidth}{!}{
\begin{tabular}{c|c|ccccc|c}
\toprule
\begin{tabular}{@{}c@{}} \textbf{Model} \\  \textbf{(Input length) }\end{tabular}  & \begin{tabular}{@{}c@{}} \textbf{SNN} \\   \end{tabular} & \begin{tabular}{@{}c@{}} \textbf{ListOps} \\ \textbf{(2,048)} \end{tabular} & \begin{tabular}{@{}c@{}} \textbf{Text} \\ \textbf{(4,096)} \end{tabular} & \begin{tabular}{@{}c@{}} \textbf{Retrieval} \\ \textbf{(4,000)} \end{tabular} & \begin{tabular}{@{}c@{}} \textbf{Image} \\ \textbf{(1,024)} \end{tabular} & \begin{tabular}{@{}c@{}} \textbf{Pathfinder} \\ \textbf{(1,024)} \end{tabular} & \textbf{Avg.} \\ \midrule
Random  & - & 10.00 & 50.00 & 50.00 & 10.00 & 50.00 & 34.00 \\
Transformer~\cite{vaswani2017attention} & \XSolidBrush & 36.37 & 64.27 & 57.46 & 42.44 & 71.40 & 54.39 \\ 
S4 (Bidirectional)~\cite{gu2021efficiently} & \XSolidBrush & 59.60 & 86.82 & 90.90 & 88.65 & 94.20 & 84.03  \\ \midrule
Binary S4D~\cite{stan2024learning} & \Checkmark & 54.80 & 82.50 & 85.03 & 82.00 & 79.79 & 77.39 \\ 
$\rightarrow$ + GSU \& GeLU & \Checkmark & 59.60 & 86.50 & 90.22 & 85.00 & 91.30 & 82.52 \\ 
SpikingSSMs~\cite{shen2025spikingssms} & \Checkmark & 60.23 & 80.41 & 88.77 & 88.21 & 93.51 & 82.23 \\
Spiking LMU~\cite{liu2024lmuformer} & \Checkmark & 37.30 & 65.80 & 79.76 & 55.65 & 72.68 & 62.23 \\ 
ELM Neuron~\cite{spieler2023expressive} & \Checkmark & 44.55 & 75.40 & 84.93 & 49.62 & 71.15 & 69.25 \\ 
SD-TCM~\cite{huang2024prf} & \Checkmark &  59.20 & 86.33 & 89.88 & 84.77 & 91.76 & 82.39 \\
\textbf{Ours} & \Checkmark & \textbf{60.02} & \textbf{86.52} & \textbf{90.02} & \textbf{85.32} & \textbf{92.36} & \textbf{82.88} \\
\bottomrule
\end{tabular}
}
\label{table:performance_comparison_lra}
\end{table}

\subsection{Sparser Spike with Accelerated Training}
To evaluate the sparsity of the D-RF method, we compare the spike firing rates and theoretical energy consumption~\cite{rathi2020diet} with those of similar approaches on the LRA dataset~\cite{tay2020long}. 
As shown in Table~\ref{compare}, our model demonstrates a significant advantage.  Specifically, on the ListOps task~\cite{nangia2018listops}, it achieves a Spiking Rate of 9.8\% and an energy consumption of 62.48mJ. Moreover, on the Image tasks~\cite{krizhevsky2009learning}, the spike firing rate is reduced by 49.7\% compared to the SD-TCM method. These results highlight the effectiveness of the adaptive threshold mechanism in enhancing energy efficiency.

\begin{figure}[htpb] 
\centering
\includegraphics[width=1.0\linewidth]{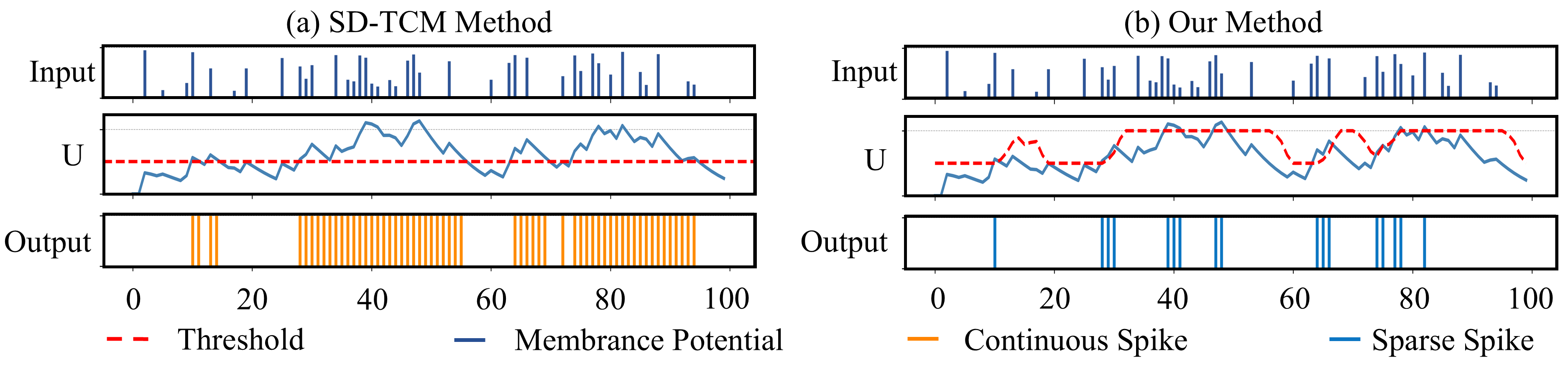}
\caption{Comparison of Spiking Behavior Across Different Methods: (a) The SD-TCM method exhibits continuous spiking activity, as reflected in its membrane potential and spike output. (b) The D-RF method shows sparser spike generation, indicating more efficient spiking behavior.}
\label{experiment}
\end{figure}

%
Additionally, we visualize the spiking behaviors of different reset methods to further the sparsity of our method. The SD-TCM model~\cite{huang2024prf} assumes that a learnable decay constant can effectively approximate the reset process, with a static threshold. In contrast, our method dynamically adjusts the threshold based on previous timesteps. As shown in Fig.~\ref{experiment}, the proposed method significantly increases spike sparsity, indicating greater potential for energy efficiency.

\begin{figure}[htpb] 
\centering
\includegraphics[scale=0.39]{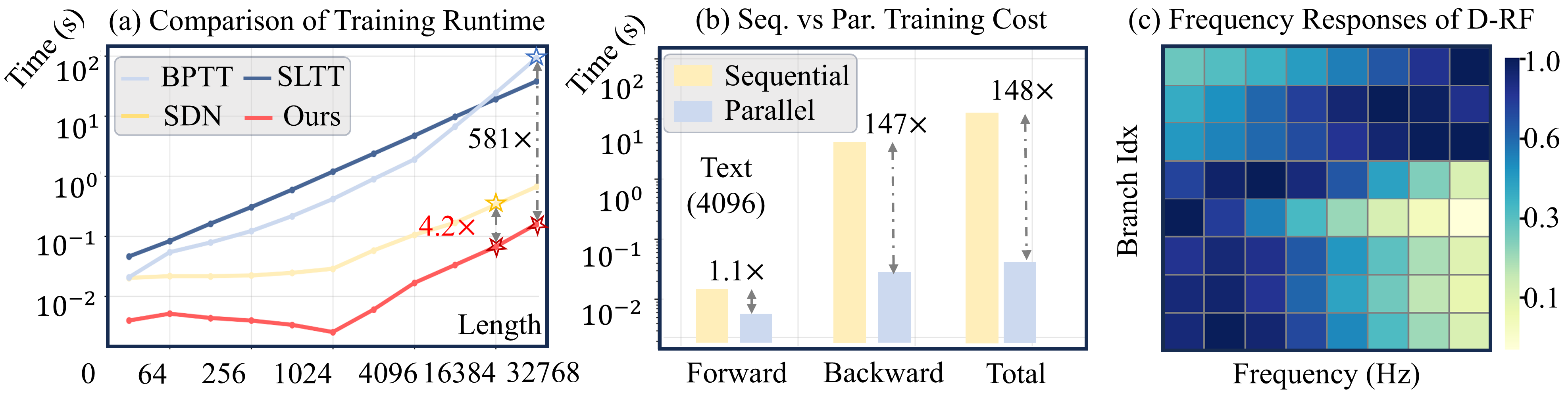}
\caption{Train Speed and Frequency Analysis of D-RF: (a) Comparison of Training Runtime. (b) Sequential vs. Parallel Training Cost: Our method accelerates both forward and backward propagation. (c) Frequency Responses of D-RF: D-RF exhibits a broader frequency response.}
\label{experiment_2}
\end{figure}

\begin{table}[]
\caption{Comparison of Metrics across the LRA Benchmark.}
\resizebox{\textwidth}{!}{
\begin{tabular}{c|c|ccccc|c}
\toprule
\textbf{Metric}  & \textbf{Method}  & \textbf{ListOps} & \textbf{Text} & \textbf{Retreival} & \textbf{Image} & \textbf{Pathfinder} & \textbf{Avg.} \\ \midrule
\multirow{3}{*}{Spiking Rate (\%)} & SpikingSSM~\cite{shen2025spikingssms} & 13.2    & 10.1 & 6.9      & 22.1  & 7.4     &   11.9   \\
                             & SD-TCM~\cite{huang2024prf}$^\dagger$ &    11.2     &   7.9   &     5.7      &  15.7     &   5.8   & 9.3  \\
                             & \textbf{Ours}          & \textbf{9.8}      &  \textbf{6.3}    &  \textbf{3.3}    &    \textbf{7.9 }  &   \textbf{3.2}   & \textbf{6.1}    \\ \midrule
\multirow{3}{*}{Energy Cost (mJ)} & SpikingSSM~\cite{shen2025spikingssms} & 84.2   &   355.2   & 237.0          & 708.9      &  65.1          &  290.1   \\
                             & SD-TCM~\cite{huang2024prf}$^\dagger$  &   71.4      &    277.8  &  195.7    & 503.6      &   51.0    & 220.6    \\
                             & \textbf{Ours} & \textbf{62.5}        &  \textbf{221.5}    &  \textbf{113.3}     &  \textbf{253.4}     &  \textbf{28.1}        &   \textbf{135.8}  \\ \bottomrule
\end{tabular}
}
\label{energy_cost}
\begin{flushleft}
  \footnotesize$^\dagger$ Results reproduced by ourselves, as the original code is not publicly available.
\end{flushleft}
\label{compare}
\end{table}
Our method also ensures high training efficiency. We compare the per-epoch training cost across different sequence lengths. As shown in Fig.~\ref{experiment_2}(a), we visualize the average epoch time under a batch size of 128. The results indicate that the proposed method achieves higher execution efficiency as the sequence length increases. At a sequence length of 32768, our method achieves a 581\(\times\) speedup over BPTT~\cite{wu2018spatio}. Furthermore, compared to the SDN method~\cite{shen2025spikingssms}, our method also shows competitive performance, achieving 4.2\(\times\) acceleration. 
We further validate the strategy on Text tasks (4096). As shown in Fig.~\ref{experiment_2}(b), benefiting from highly parallelized membrane potential accumulation and spike generation processes, D-RF achieves faster simulation on GPU, yielding a 1.1 \(\times\) speedup. Moreover, the adaptive threshold mechanism enables the transformation of the serial accumulation-decay-firing process into a parallelizable form, resulting in up to 147\(\times\) training acceleration. These results confirm that D-RF effectively addresses the high training cost. 

\subsection{Ablation Experiment}
\begin{wraptable}{r}{0.57\textwidth}
    \vspace{-12mm} 
    \centering
    \caption{Ablation Experiment}
    \label{tab:number}
    \begin{tabular}{cccccc}
        \toprule
        \multirow{2}{*}{Dataset}   & \multirow{2}{*}{Method} & \multicolumn{4}{c}{Number of Dendrites} \\ \cmidrule{3-6} 
                                   &                         & n=1       & n=4      & n=8      & n=16     \\ \midrule
        \multirow{2}{*}{S-CIFAR10} & adaptive                & 80.3      & 84.3     & 84.6     & \textbf{85.1}    \\
                                   & w/o                     & 79.2      &   83.9       &   84.1       &    84.9      \\ \midrule
        \multirow{2}{*}{ListOps}   & adaptive                & 55.2      & 59.1     & 60.2     & \textbf{60.3}     \\
                                   & w/o                     & 54.2      & 58.9     & 59.2     & 59.6     \\ \bottomrule
    \end{tabular}
\end{wraptable}
To assess the impact of the dendritic structure and the adaptive threshold mechanism, we conduct ablation experiments on the S-CIFAR10 and ListOps datasets. We compare the performance in different numbers of dendrites (n = 1, 4, 8, 16). As shown in Table~\ref{tab:number}, the model performance improves with an increasing number of dendrites. 
Considering the trade-off between complexity and performance, we set the number of dendrites to $n = 4$ for the S-CIFAR10 dataset and $n = 8$ for the LRA dataset. As shown in Fig.~\ref{experiment}(c), we visualize the frequency responses of individual dendritic branches. The results indicate that the proposed D-RF neuron captures nearly the entire frequency spectrum, further confirming the effectiveness of the dendritic structure. In addition, we evaluate the effectiveness of the adaptive threshold mechanism. Experimental results demonstrate that the adaptive threshold improves model performance, primarily by effectively suppressing the adverse impact of redundant feature information on the results.

\section{Conclusion}
Inspired by the dendritic structure of biological neurons, this study proposes the D-RF model to further enhance the performance of SNNs on time-series signals. It consists of a multi-dendritic and soma structure. The multi-dendritic structure consists of branches with distinct decay factors, enabling the neuron to effectively extract multi-frequency information from the input signal. The soma with an adaptive threshold that ensures sparse spiking while enabling parallelizable computation. 
Extensive experiments demonstrate that our model achieves competitive results while maintaining sparse spiking activity for efficient training. These results highlight the strong potential of D-RF method to enable effective and efficient long sequence modeling on edge-computing platforms.

\bibliographystyle{plain}

\bibliography{neurips_2025}

\newpage
\end{document}